\documentclass{article}

\usepackage{arxiv}

\usepackage[utf8]{inputenc} 
\usepackage[T1]{fontenc}    
\usepackage{hyperref}       
\usepackage{url}            
\usepackage{booktabs}       
\usepackage{amsfonts}       
\usepackage{nicefrac}       
\usepackage{microtype}      
\usepackage{graphicx}
\usepackage{float}

\title{Effects of padding on LSTMs and CNNs}

\author{
  Dwarampudi Mahidhar Reddy\\
  Computer Science and Engineering Department\\
  Manipal Institute of Technology\\
  Manipal, Karnataka, 576104, India \\
  \texttt{mahidhar\_d@hotmail.com} \\
   \And
  N V Subba Reddy \\
  Computer Science and Engineering Department\\
  Manipal Institute of Technology\\
  Manipal, Karnataka, 576104, India \\
  \texttt{nvs.reddy@manipal.edu} \\
}

\begin{document}
\maketitle

\begin{abstract}
Long Short-Term Memory (LSTM) Networks and Convolutional Neural Networks (CNN) have become very common and are used in many fields as they were effective in solving many problems where the general neural networks were inefficient. They were applied to various problems mostly related to images and sequences. Since LSTMs and CNNs take inputs of same length and dimension, input images and sequences are padded to maximum length while testing and training. This padding can affect the way the networks functions and can make a great deal when it comes to performance and accuracies. This paper studies this and suggests the best way to pad an input sequence. This paper uses a simple sentiment analysis task for this purpose. We use the same dataset on both the networks with various padding to show the difference. This paper also discusses some preprocessing techniques done on the data to ensure effective analysis of the data.
\end{abstract}


\section{Introduction}
Long Shot Term Memory (LSTM) Networks and Convolutional Neural Networks(CNNs) are used in various fields. LSTM and CNN take sequential inputs of equal length. Hence, all the inputs should be padded to make the lengths of the inputs equal. This paper considers a common task for both CNN and LSTM and analyses the effect of padding on them, the task being Sentiment Analysis. We study 2 types of padding, namely pre and post padding. We use twitter data to classify the tweets, into 2 sentiments, positive and negative. We preprocess the data and take word vectors, or distributed representation of words for this purpose.

\section{Literature Survey}
LSTMs are being used in many areas these days such as, Machine Translation \cite{s2s} as shown in “Sequence to Sequence Learning with Neural Networks” by Ilya Sutskever, Oriol Vinyals and Quoc V. Le, Image Captioning\cite{2} as shown in “Show and tell: A neural image caption generator” by Vinyals, Alexander and Samy, Hand writing generation\cite{3} in “Generating Sequences With Recurent Neural Networks”  by Alex Graves and Question answering system\cite{4} by Di Wang and Eric Nyberg in “A Long Short-Term Memory Model for Answer Sentence Selection in Question Answering”. And CNNs are mostly used for pattern recognition either in images or in text. It was highly used for image classification one such example is ImageNet\cite{5} by Alex Krizhevsky, Ilya Sutskever and Geoffrey E. Hinton. It was also highly used for pattern recognition tasks such as facial recognition\cite{6}, as shown in “Face recognition: a convolutional neural-network approach” by S. Lawrence, C.L. Giles, Ah Chung Tsoi and A. D. Black. Until recently when people started using it for Natural Language Processing tasks like, Sentence Classification\cite{7} in “Convolutional Neural Networks for Sentence Classification” by Yoon Kim and Sentiment Analysis\cite{8}, as shown in “Deep Convolutional Neural Networks for Sentiment Analysis of Short Texts” by Cicero Nogueira dos Santos and Maira Gatti.
\par
In all these examples the sequences were padded to maximum length in order to train and test them on LSTMs and CNNs. Here, we see various ways of preprocessing data for variable length inputs. Then a brief discussion of word vectors. Then we talk about LSTMs and CNNs. 

\subsection{Preprocessing Variable Length Input Sequences}
\begin{figure}[H]
	\centering
	\fbox{ \includegraphics[scale=0.25]{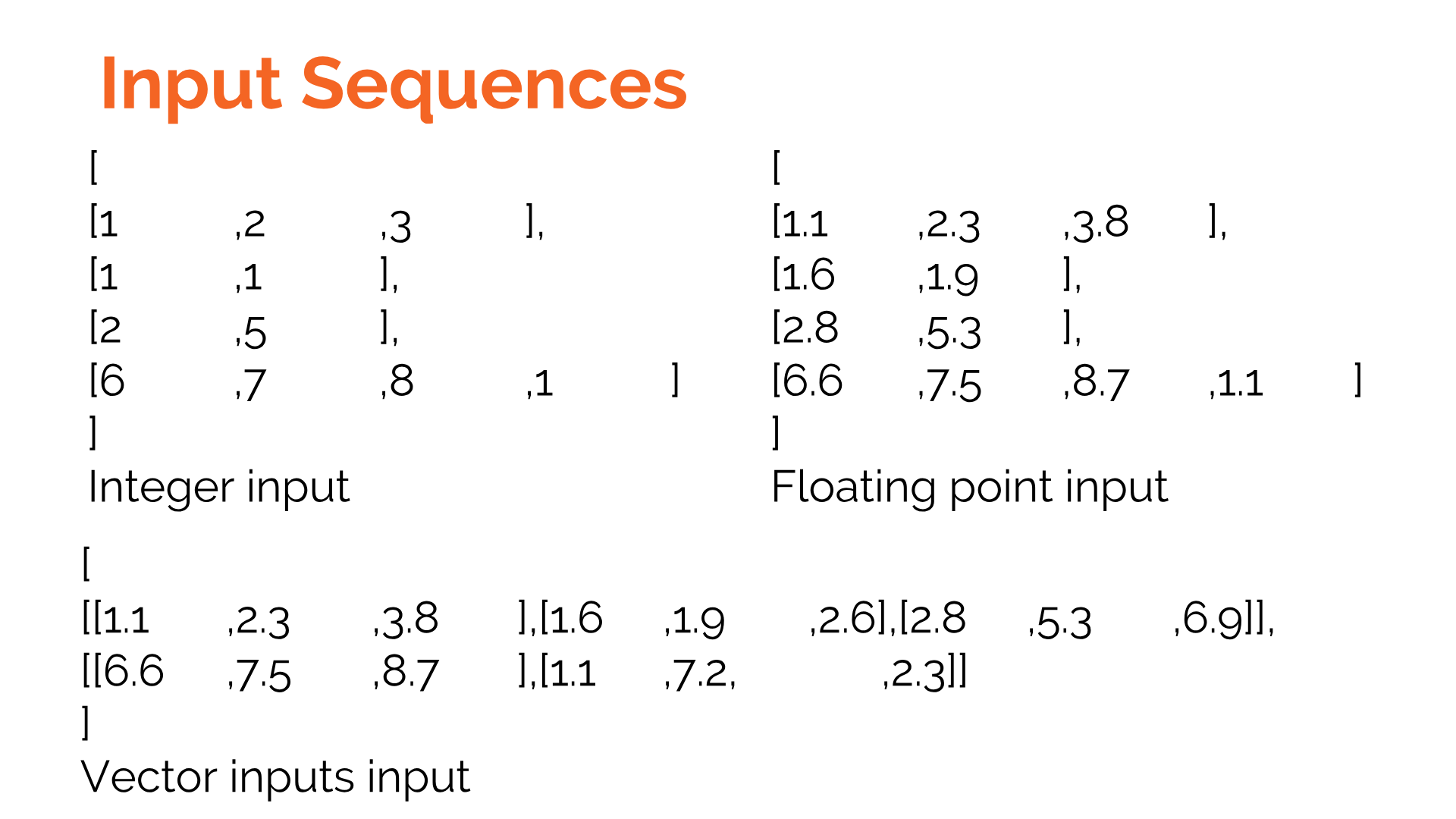}}
	\caption{Examples of input sequences}
	\label{fig:input_sequences}
\end{figure}

Generally, we have sequences of integers, floating points or vectors of the afore mentioned as inputs. These sequences are of variable lengths and not constant lengths, Figure \ref{fig:input_sequences}.

We have to make them of equal length before we can use them for training. There are two ways of doing this:

\subsubsection{Truncation}
\begin{itemize}
	\item Pre-Sequence Truncation: All the sequences are truncated in the beginning according to the length of the smallest sequence or a chosen length. (Figure \ref{fig:pre_truncate}). If the chosen length is longer than the smallest sequence, the smallest sequence is padded accordingly.
	\begin{figure}[H]
		\centering
		\fbox{ \includegraphics[scale=0.25]{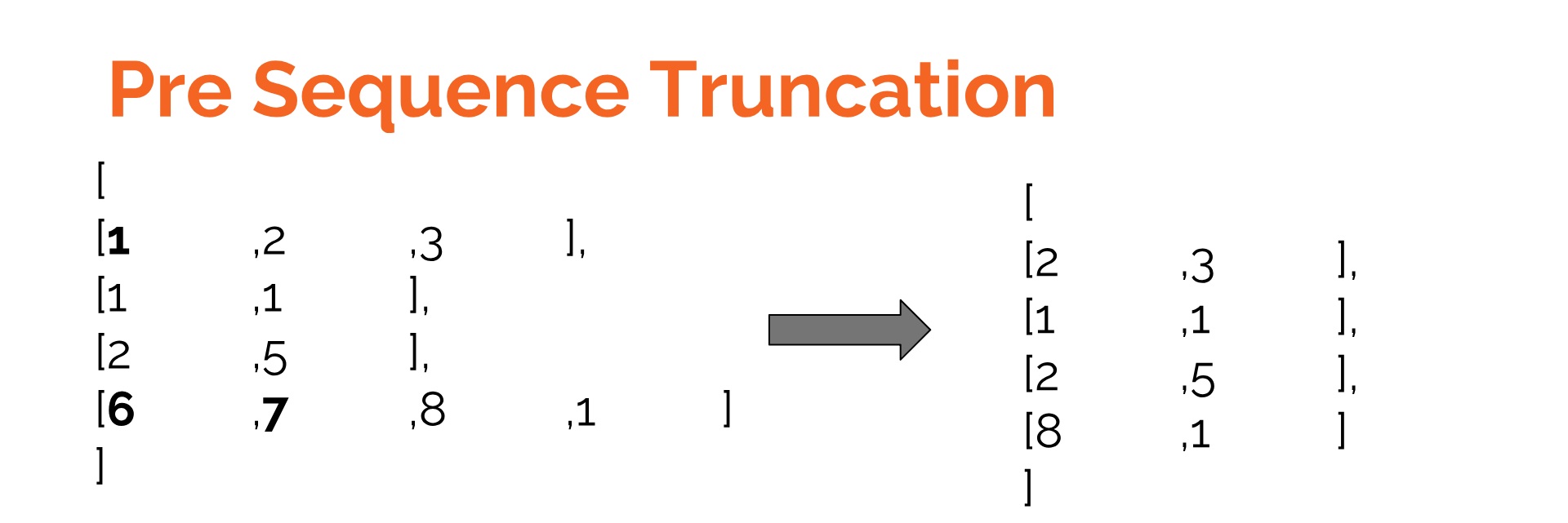}}
		\caption{Pre-Sequence Truncation, the values in bold are removed after truncation}
		\label{fig:pre_truncate}
	\end{figure}
	
	\item Post-Sequence Truncation: All the sequences are truncated in the ending according to the length of the smallest sequence or a chosen length. (Figure \ref{fig:post_truncate}). If the chosen length is longer than the smallest sequence, the smallest sequence is padded accordingly.
	\begin{figure}[H]
		\centering
		\fbox{ \includegraphics[scale=0.25]{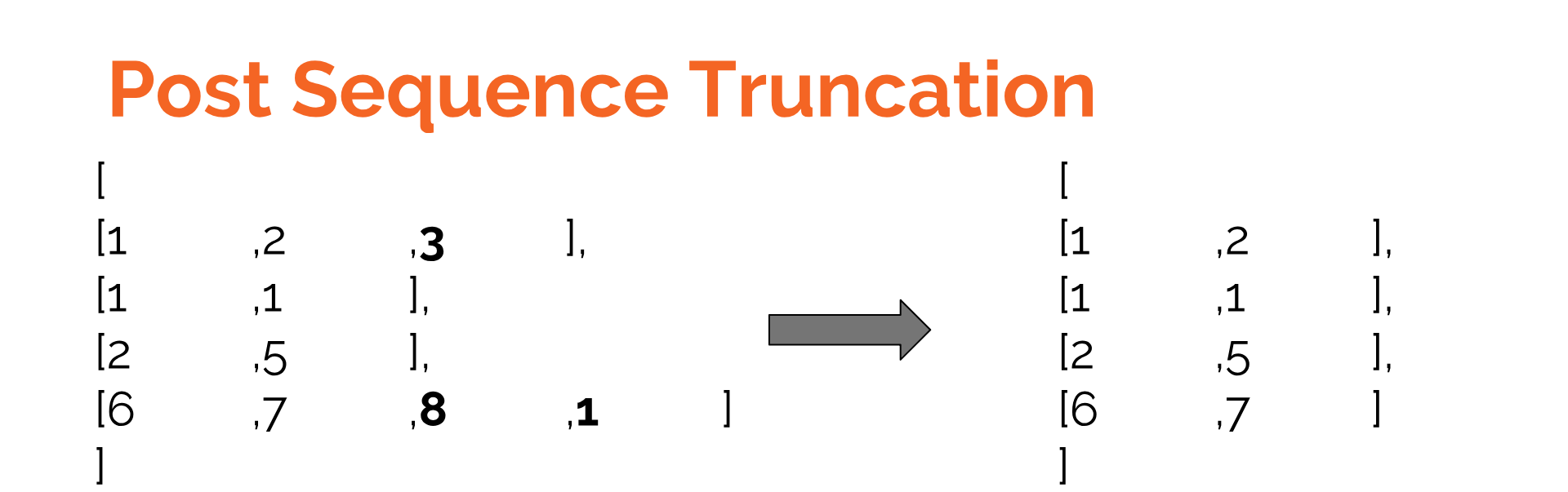}}
		\caption{Post Sequence Padding. The values in bold are removed after truncation.}
		\label{fig:post_truncate}
	\end{figure}
\end{itemize}

\subsubsection{Padding}
\begin{itemize}
	\item Pre-Padding: All the sequences are padded with zeroes in the beginning according to the longest sequence’s length or a chosen length longer than the longest length. (Figure \ref{fig:pre_padding})
	\begin{figure}[H]
		\centering
		\fbox{ \includegraphics[scale=0.25]{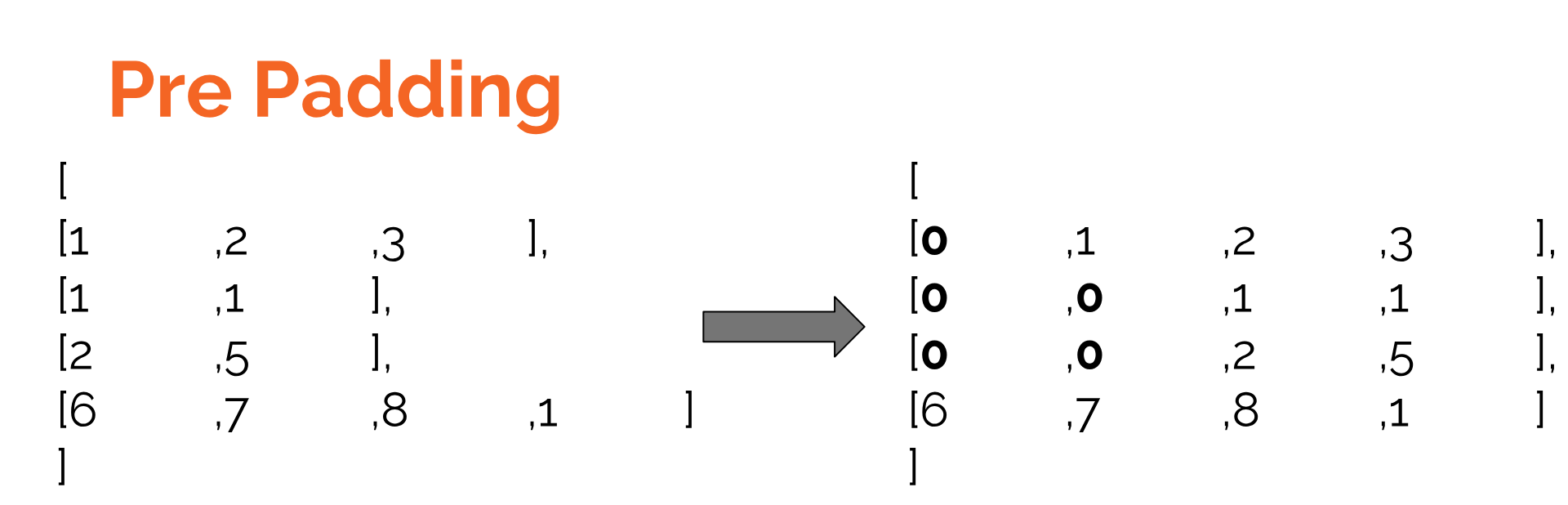}}
		\caption{Pre-Padding, the zeroes in bold are added after padding}
		\label{fig:pre_padding}
	\end{figure}
	
	\item Post-Padding: All the sequences are padded with zeroes in the ending according to the longest sequence’s length or a chosen length longer than the longest length. (Figure \ref{fig:post_padding})
	\begin{figure}[H]
		\centering
		\fbox{ \includegraphics[scale=0.25]{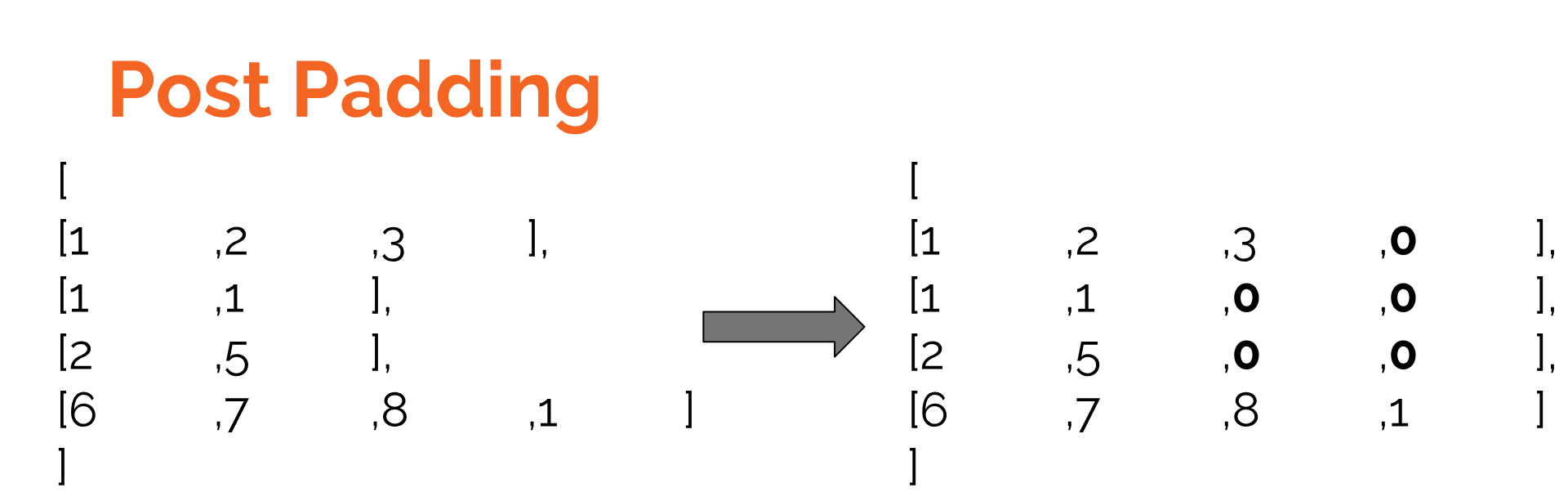}}
		\caption{Post Padding, the zeros in bold are added after padding.}
		\label{fig:post_padding}
	\end{figure}
\end{itemize}

As we can see from Figure \ref{fig:pre_truncate} and \ref{fig:post_truncate} that there is a clear data loss in truncating, truncation is not a viable option for converting the sequences to equal length. So, we prefer padding over and truncating. Here, we compare post padding to pre-padding on LSTMs and CNNs.

\subsection{Word Vectors}
Instead of using Bag of Words Representation (BOW) or Continuous Bag of Words (CBOW), we are using continuous bag of word vectors representation. Where we take sequence of word in the order of their appearance, as input for the model.
\par
Here, we are using Word2Vec skipgram\cite{9}, for building the word vectors. Word Vectors in skipgram model are constructed by considering a fake task, which is predict the window of words around the given words. Usually the neural network contains all the vocabulary in the input layer and the output layer and one hidden layer, on which the number of nodes will define the dimensions of the word vectors. We first consider a word and a window around it, a pair of the given word and every word in its window becomes a training vector. After a few epochs we remove the last layer and take the layer before it or the penultimate layer into consideration whose output will be our word vector for a given activated word in the input layer.

\subsection{LSTMs and CNNs}
LSTMs\cite{10} are a modification of recurrent neural networks (RNNs)\cite{11}. RNNs consider their previous output as an input along with their next input, this allows them keep track of their previous output making them good to work with sequences. The slight modification of LSTMs is they have memory or “cell state” which keeps updating with the sequence, giving them more information making them very good while working with sequences, where time plays an important role, like time series forecasting. 
\par
CNNs\cite{12} on the other hand doesn’t have memory but instead tries to find pattern in the given data. Neurons in CNNs unlike normal neural networks have learnable weights and biases. These neurons receive multiple inputs and take a weighted sum over them before passing them through an activation function, which will throw an output. In addition to this CNNs input is multi channeled. CNNs have filters which slide over the input data take dot product and ads a bias before throwing a number.
\par
These filters in further layers extract very high-level features and create multiple feature maps. In CNNs the neurons are not fully connected like the normal neural network instead they are only connected to a subset of input data. This reduces the number of parameter in the whole network. Next comes a pooling layer which reduces the spatial representation to reduce the number of parameters and computation in the network. All the neural networks end with a fully connected regular neural network which essentially does the classification for the CNN.

\section{Data}
The dataset used in this paper was published on the Thinknook website \cite{13}. Only 10\% of all the tweets were taken in this paper. Around 157,860 tweets were taken and divided into train test data, the number of train tweets is 126,288 tweets consisting of 63,001 positive and 63,287 negative tweets. The test data consisted of 31572 tweets. All the numbers in these tweets were replaced by ‘0’, twitter handles by ‘1’ and URLs by ‘2’. All the words in the text were lemmatized \cite{14}, stop words were not removed as negative stopwords can affect sentiment of the tweet \cite{15}. These were then trained on the Word2Vec skipgram model, with a hidden of 100 units for 5 epochs, creating word vectors of size 100 after the training. The window size was 5. The longest tweet was of size 93. So all the sequences were padded to that length.

\section{Final Comparison}
The LSTM used in this paper has 4 hidden layers. Each layer has 100 neurons. A dropout of 0.2 was used on each layer with an additional recurrent dropout of 0.2. The LSTM was set to take sequences of length 93 as that was the maximum length of the tweet. Tanh was used as the activation function for all the layers except the last layer (output layer) where sigmoid was used as activation for classification.
\par
The CNN used in this paper are similar to the LSTMs used, they have 4 hidden layers of 100 units each with a dropout of 0.2 on each layer. Linear activation function was used on each layer except the output layer where a fully connected neuron was used for prediction with sigmoid activation function.

\section{Results}
The following tables compare the models, padding and their accuracies.

\begin{table}[H]
	\caption{LSTM pre-padding vs LSTM post-padding (\%)}
	\centering
	\begin{tabular}{lll}
		\toprule

		     & LSTM-4 Pre-Padding     & LSTM-4 Post-Padding \\
		\midrule
		Train & 80.072  & 49.977    \\
		Test     & 80.321 & 50.117      \\
		Epochs     & 9       & 6  \\
		\bottomrule
	\end{tabular}
	\label{tab:lstm}
\end{table}

Though post padding model peaked it's efficiency at 6 epochs and started to overfit after that, it’s accuracy is way less than pre-padding. Artificial Neural Networks are inspired from biological neural networks. This can be compared to the way people talk. All the zeroes are silence. Say, a person X is talking to person Y, person X waits for some time and talks, Y immediately replies, there’s a higher chance that Y remembers most of what X said and Y’s reply will be more relevant to what X talked about. Say, X said something to Y and Y didn’t reply immediately, but took some time, the longer Y waits to reply, the more context Y looses of the conversation, and more irrelevant his reply to X will be.
\par
\begin{table}[H]
	\caption{CNN pre-padding vs post padding. (\%)}
	\centering
	\begin{tabular}{lll}
		\toprule
		
		& CNN-4 Pre-Padding     & CNN-4 Post-Padding \\
		\midrule
		Train & 74.721  & 74.465    \\
		Test     & 74.994 & 74.908      \\
		Epochs     & 10       & 10  \\
		\bottomrule
	\end{tabular}
	\label{tab:cnn}
\end{table}

Pre-padding and post padding doesn’t matter much to CNN because unlike LSTMs, CNNs don’t try to remember stuff from the previous output, but instead tries to find pattern in the given data. For Sentiment Analysis, LSTMs are more efficient than CNNs. But it is better to pre- pad sequences as they were more efficient in case of LSTM. In general, pre-padding would be better when multiple types of neural networks are combined to perform a task.

\bibliographystyle{unsrt}

\end{document}